\documentclass[letterpaper, 10 pt, conference]{ieeeconf}  

\IEEEoverridecommandlockouts                              

\usepackage[acronym]{glossaries}
\usepackage[utf8]{inputenc}
\usepackage{amsfonts}
\usepackage{amsmath}
\usepackage{amssymb}
\usepackage{booktabs}
\usepackage{graphicx}
\usepackage{hyperref}
\usepackage{multicol}
\usepackage{multirow}
\usepackage{siunitx}
\usepackage{xcolor}
\usepackage{multirow}

\usepackage{fourier} 
\usepackage{array}
\usepackage{makecell}

\usepackage{caption}
\newcommand{\acronym}[1]{\gls{#1}\@}
\newcommand{\acronympl}[1]{\glspl{#1}\@}

\newacronym{fpfh}{FPFH}{Fast Point Feature Histogram}
\newacronym{icp}{ICP}{Iterative Closest Point}
\newacronym{iss}{ISS}{Intrinsic Shape Signature}
\newacronym{pca}{PCA}{Principal Component Analysis}
\newacronym{csg}{CSG}{Constrained Similarity Graph}
\newacronym{crf}{CRF}{Conditional Random Field}
\newacronym{board}{BOARD}{BOrder Aware Repeatable Directions}
\newacronym{esdf}{ESDF}{Euclidean Signed Distance Field}
\newacronym{tsdf}{TSDF}{Truncated Signed Distance Field}
\newacronym{ros}{ROS}{Robot Operating System}
\newacronym{gsm}{GSM}{Global Segmentation Map}
\newacronym{slam}{SLAM}{Simultaneous Localization and Mapping}
\newacronym{rmse}{RMSE}{Root-mean-square error}
\newacronym{ransac}{RANSAC}{Random Sample Consensus}
\newacronym{mav}{MAV}{Micro Air Vehicle}
\newacronym{uav}{UAV}{Unmanned Aerial Vehicle}
\newacronym{mavs}{MAVs}{Micro Air Vehicles}
\newacronym{uavs}{UAVs}{Unmanned Aerial Vehicles}
\newacronym{vio}{VIO}{Visual-Inertial Odometry}
\newacronym{viwls}{VIWLS}{Visual Inertial Weighted Least-Squares}
\newacronym{msf}{MSF}{Multi Sensor Fusion}
\newacronym{imu}{IMU}{Inertial Measurement Unit}
\newacronym{tnr}{T\&R}{Teach\&Repeat}
\newacronym{mpc}{MPC}{Model Predictive Controller}
\newacronym{rel}{rel}{mean absolute relative error}
\newacronym{fp}{FP}{False Positive}
\newacronym{fn}{FN}{False Negative}
\newacronym{cnn}{CNN}{Convolutional Neural Network}

\definecolor{todo-red}{RGB}{200,12,12}
\definecolor{green4}{RGB}{0,128,0}
\definecolor{pink}{RGB}{255,20,147}

\newcommand{\vs}{\mathbf{s}}
\newcommand{\vt}{\mathbf{t}}

\newcommand{\lossone}{\mathcal{L}_1}

\title{\LARGE \bf
Predicting Unobserved Space For Planning \\ via Depth Map Augmentation
}

\author{$^*$Marius Fehr$^{1}$, $^*$Tim~Taubner$^{1}$, Yang Liu$^{1}$, Roland Siegwart$^{1}$, Cesar Cadena$^{1}$ \\
\thanks{$^{1}$Autonomous Systems Lab, ETH Z\"urich, Switzerland \newline {\tt \{fehrma, taubnert, lya, rsiegwart, cesarc\}@ethz.ch} \newline $^*$equally contributed}%
}

\begin{document}

\maketitle
\thispagestyle{empty}
\pagestyle{empty}


\begin{abstract}
%
%
%
%
%
Safe and efficient path planning is crucial for autonomous mobile robots.
A prerequisite for path planning is to have a comprehensive understanding of the 3D structure of the robot's environment.
On \acronym{mavs} this is commonly achieved using low-cost sensors, such as stereo or RGB-D cameras.
These sensors may fail to provide depth measurements in textureless or IR-absorbing areas and have limited effective range.
In path planning, this results in inefficient trajectories or failure to recognize a feasible path to the goal, hence significantly impairing the robot's mobility.
Recent advances in deep learning enables us to exploit prior experience about the shape of the world and hence to infer complete depth maps from color images and additional sparse depth measurements.
In this work, we present an augmented planning system and investigate the effects of employing state-of-the-art depth completion techniques, specifically trained to augment sparse depth maps originating from RGB-D sensors, semi-dense methods and stereo matchers.
We extensively evaluate our approach in online path planning experiments based on simulated data, as well as global path planning experiments based on real world \acronym{mav} data.
We show that our augmented system, provided with only sparse depth perception, can reach on-par performance to ground truth depth input in simulated online planning experiments.
On real world \acronym{mav} data the augmented system demonstrates superior performance compared to a planner based on very dense RGB-D depth maps.
\end{abstract}


\section{INTRODUCTION}
\label{sec:intro}
Autonomous mobile robots are deployed in challenging environments to perform inspection, mapping and surveillance tasks.
For many of these applications the environment is initially unknown and the robot must solely rely on its own perception of the world.
For \acronympl{mav} this is particularly challenging, as weight restrictions tend to prohibit the use of more powerful depth sensing modalities, such as LiDAR.
For the purpose of path planning most \acronym{mav} systems employ vision based sensors, such as stereo cameras, low-cost RGB-D sensors or (semi-) dense visual(-inertial) mapping techniques.
The common denominator of all of these methods is that the provided depth maps are potentially sparse and fail to reliably provide depth estimates for textureless or IR-absorbing surfaces, in addition to a limited \textit{reliable} range.
These missing estimates, however, make it difficult to clear the free space ahead of the \acronym{mav}.


\begin{figure}[ht]
	    \centering
		\includegraphics[width=.8\columnwidth]{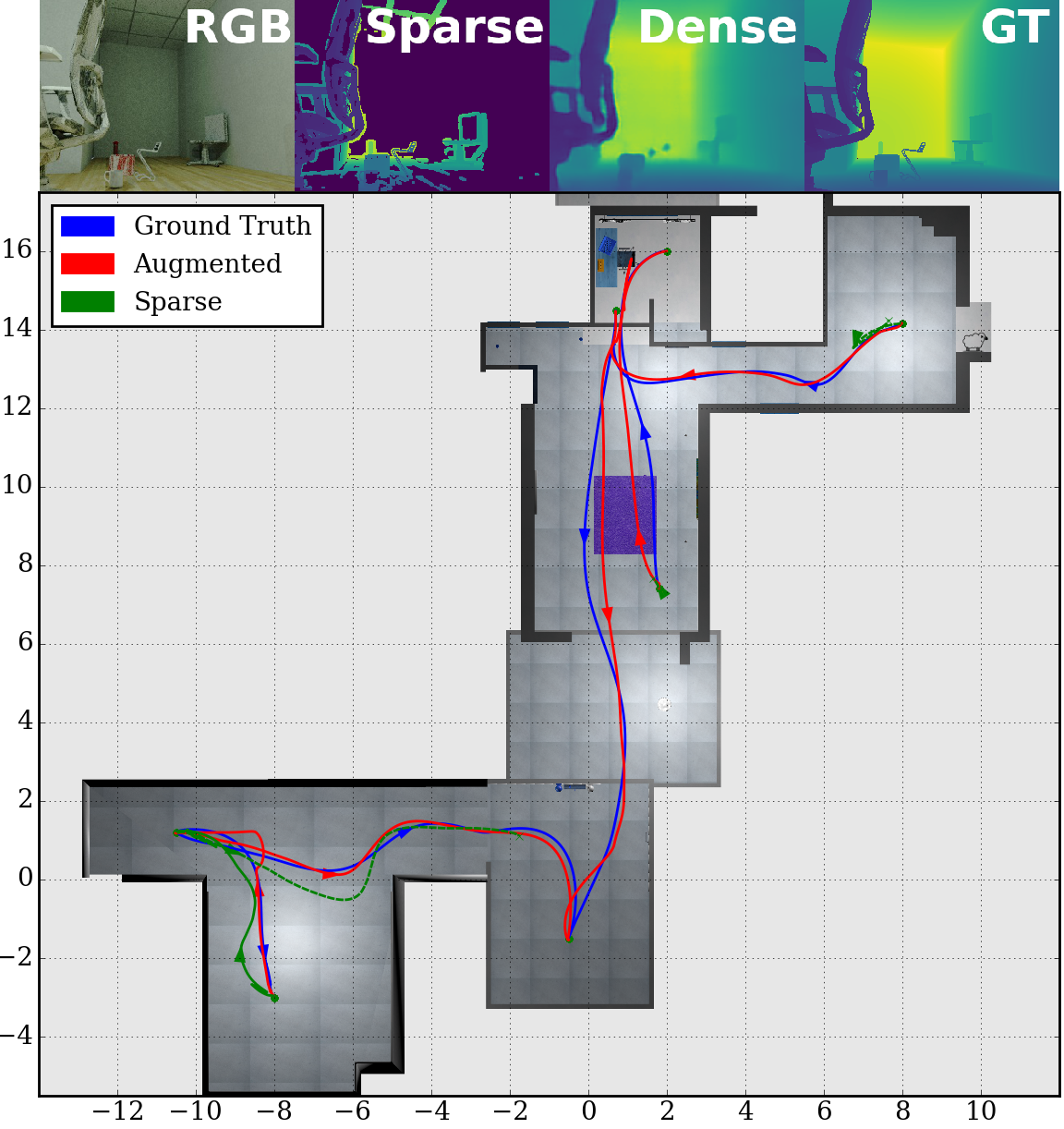} 
		\caption{
		\emph{(top)} RGB and sparse depth input and the CNN-augmented dense depth map compared to the ground truth (GT).
		\newline
		\emph{(bottom)} Paths resulting from GT, CNN-augmented and sparse depth maps.
		Dashed lines indicate failure.}
		\label{fig:traj}
\end{figure}


In \acronym{mav} planning research there are typically two different ways how planners deal with unobserved space resulting from limited depth perception.
Optimistic planners~\cite{pivtoraiko2013incremental, chen2016online} assume some of the unobserved space to be free, which works well in low-density areas, but forfeits all safety guarantees.
Conservative planners~\cite{oleynikovasafe2018} on the other hand assume unobserved space to be entirely occupied.
If the free space remains unobserved, feasible paths might get discarded in favor of highly inefficient ones, or no solution is found at all.
In this work, we aim to improve path planning by augmenting the depth perception using a CNN  based  predictor to make a more informed decision about the state of unobserved space.
Therefore -- inspired by the human vision system -- we propose a system that tackles this problem by augmenting these sparse depth maps with predictions based on the previously learned shape of the world.
We retrain the residual neural network proposed by~\cite{ma2017sparsedense} to augment sparse depth maps by predicting unknown pixels from color images.
To that end, we emulate commonly occurring failures of real depth sensors, depth estimation, and mapping techniques such as missing depth measurements in low-texture areas, to generate training data.
The resulting \textit{augmented} sensor provides complete color and depth information over its whole field-of-view.
This sensor and the reference depth sensors are then used and evaluated using a planning system consisting of voxblox~\cite{oleynikova2017voxblox}, a volumetric \acronym{tsdf}-based map representation and the planners proposed by~\cite{oleynikovasafe2018} and~\cite{richter2016polynomial}.
Note that in this work we do not focus on advancing the depth augmentation performance but instead evaluate the implications of reinforcing a path planning pipeline using depth augmentation.
To the best of our knowledge we are the first to do so.


The contributions of this work are as follows:
\begin{itemize}
 \item We propose and evaluate a path planning system with an augmented depth sensor at its core that, despite the sparse depth input, is able to reach promising planning performance when compared to the system using all available depth information.
 \item We have conducted thorough evaluations of the augmented planning system in online planning experiments in simulation as well as global path planning experiments based on real world data collected using an \acronym{mav} equipped with an RGB-D sensor.
\end{itemize}


\section{RELATED WORK}
\label{sec:rw}
In recent years, learning-based methods have become the state-of-the-art for inferring 3D structure from color and partial depth information.
In the field of path planning, methods that are able to learn a mapping directly from sensor data to controller inputs have achieved promising results.


\subsection{Inference of 3D structure}
There is a large body of work in the area of inferring semantic and depth information from color images.
Most traditional methods have been outperformed by different \acronym{cnn} architectures, such as the ones proposed and evaluated  in \cite{liu2015deep}, \cite{eigen2015predicting} and \cite{laina2016deeper}.
\cite{roy2016monocular} propose a neural regression forest for monocular depth estimation.
As the absolute scale of depth is not observable from color information alone, these network either have to infer the absolute scale (e.g.\ by learning typical sizes of objects) or provide only relative depth information.

Motivated by \cite{roy2016monocular}, \acronympl{cnn} have been extended to incorporate partial depth information.
\cite{pan2018depth} consider sparse depth information from multi-beam LiDAR reprojected into the camera frame as additional input to motion-blurred color images.
\cite{liao2017parse} employ a single-line LiDAR to provide absolute depth information.
\cite{ma2017sparsedense} propose a neural network using sparse depth information uniformly distributed over the image capable of inferring complete depth maps with high precision.
\cite{martins2018fusion} tackles the challenge of obtaining adequate training data by proposing a method for self-supervised learning of depth estimates from monocular images that are then fused with depth maps obtained from stereo.
%

Triggered by the KITTI depth completion benchmark the following depth augmentation methods have been proposed:
\cite{jaritz2018sparse} explicitly tackle the problem of sparsity in depth maps originating from different depth sensing modalities and employ a convolutional encoder-decoder architecture to predict depth or semantic labels.
\cite{pilzer2018unsupervised} propose another unsupervised learning method to overcome the lack of suitable training data by employing an unsupervised adversarial learning framework for depth augmentation.
%
\cite{zhang2018deep} focuses as well on solving the problem of augmenting real depth sensor measurements.
Their results demonstrate that learning to predict surface normals and occlusion boundaries from RGB combined with an optimization is able to significantly outperform pure learning-based depth augmentation methods.
%
%
Unfortunately, their solution is prohibitively expensive for \acronym{mav} adoption\footnote{Inference takes 2\,s/frame on a Xeon PC equipped with a TitanX GPU.}.
Similar to the network proposed by~\cite{ma2017sparsedense}, which was used by the proposed system, \cite{cheng2018depth} complete depth maps from uniform depth samples and color images but achieve sharper and more accurate results by employing a convolutional spatial propagation network to learn the affinity matrix.
Recent work by \cite{ma2018self} proposes a self-supervised depth completion technique that leverages sparse LiDAR data and color images and therefore does not require semi-dense depth annotation.
Using the same input data \cite{imran2019depth} focuses on preventing depth smearing between objects by introducing a novel depth representation called Depth Coefficients.
We believe that any advance on this topic that show real-time capabilities on resource constraint platforms can be easily adopted by the proposed system and serve as an alternate depth completion architecture in the future.

Instead of augmenting depth maps another option would be to infer depth information directly in the 3D map.
\cite{yang20173d} generalizes residual networks to infer complete 3D object shapes from single view depth images, an interesting approach as it bridges 2D and 3D depth/shape completion.
A fully three dimensional multi-modal network proposed in \cite{dai2017scancomplete} is capable of completing a partial 3D occupancy grid while providing per-voxel semantic information. 
An iterative scheme allows a trade-off between accuracy and speed.
However, as these approaches do not yet run in real-time they are not feasible for use in online planning onboard an \acronym{mav}.


\subsection{Learning-based end-to-end navigation}
In contrast to combining depth map augmentation with more traditional planning algorithms, as proposed in this work, there have been a variety of approaches that propose to learn robot navigation end-to-end.
In these systems, the neural network essentially takes the role of a human pilot by learning controller inputs directly from color and/or depth images.
For \acronym{mav}s this has been successfully employed in indoor \cite{kim2015deep} and outdoor scenes such as forests \cite{ross2013learning}.
\cite{loquercio2018dronet} proposed an eight layer residual network trained on data captured from cars to command a \acronym{mav} flying in outdoor street scenes.
Letting a neural network command the vehicle directly poses a certain risk as there is no guarantee that the network will perform sensible in unknown environments.
\cite{richtersafe2017} addresses this issue partially in a race-car setting.
In familiar environments a neural network allows the car to navigate significantly faster than a conservative policy would allow.
Novelty detection reactivates a safe fallback policy in unknown environments.

End-to-end learning approaches do not build global maps.
Instead, they rely on local decisions making them prone to get trapped in a local minimum.
These limited planning capabilities make them unsuitable to navigate in environments with complex topologies.
In contrast, our approach combines the ability of neural networks to infer complex shapes with a traditional planner able to overcome these local minima.


\section{AUGMENTED PLANNING SYSTEM}
\label{sec:pp}
\begin{figure}[ht]
    \centering
    \includegraphics[width=.8\columnwidth]{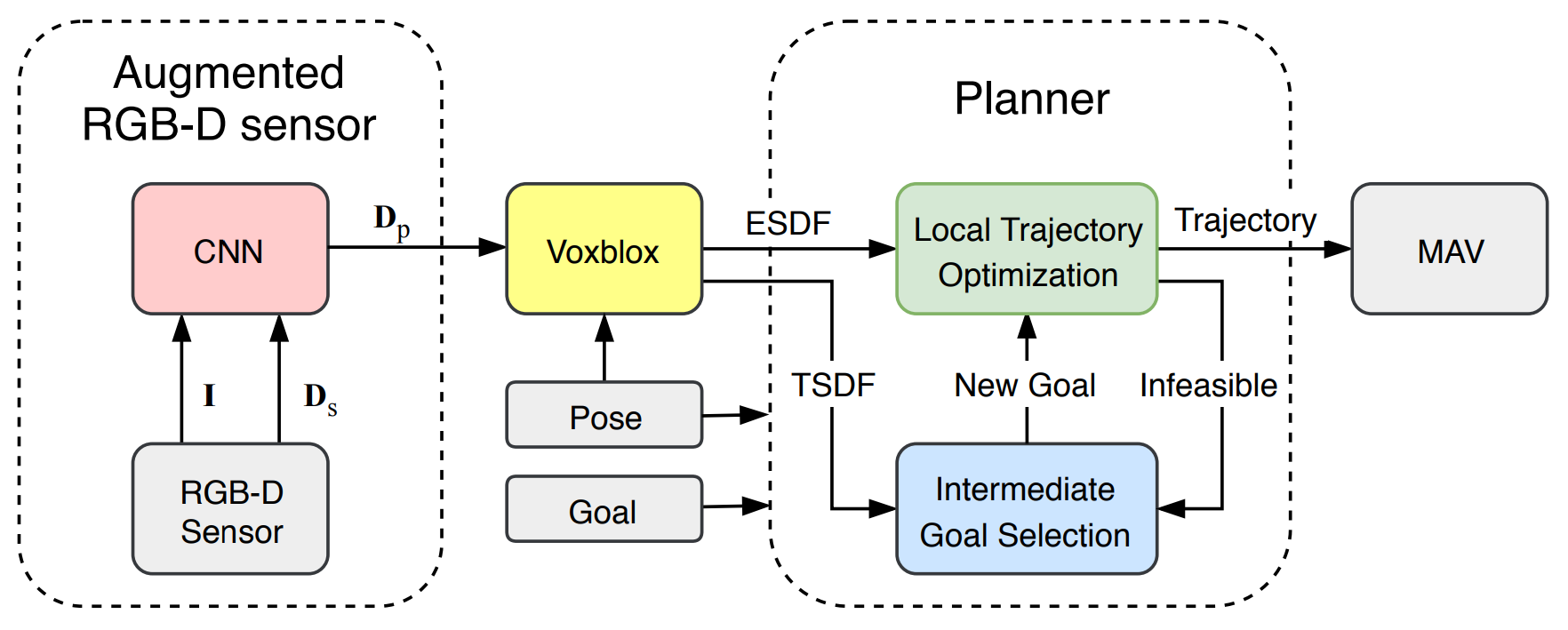}
    \caption{Pipeline consisting of enhanced RGB-D sensor, incremental map builder~\cite{oleynikova2017voxblox} and planner~\cite{oleynikovasafe2018} commanding an \acronym{mav}.}
    \label{fig:pipe}
\end{figure}

A traditional planning pipeline typically consists of an input depth sensor, map building and the planning method itself.
We propose to use a \acronym{cnn} based predictor on top of the depth sensor and treat it as a single augmented RGB-D sensor capable of sensing dense color and depth information over a common field-of-view.
The augmented depth is fused into a \acronym{tsdf} grid and an incremental \acronym{esdf} map builder provided by voxblox~\cite{oleynikova2017voxblox} is used to obtain the map representation.
The continuously updated map is then fed into a state-of-the-art, conservative path planner~\cite{oleynikovasafe2018}.
In the experiment on real \acronym{mav} data we furthermore deploy and compare the planner proposed by~\cite{richter2016polynomial}.
An overview of the pipeline can be found in Fig.~\ref{fig:pipe}.
%


\subsection{Depth Map Augmentation}
\label{sec:pp:dm}
%
%



We retrain a depth augmentation method based on the residual neural network proposed by \cite{ma2017sparsedense} motivated by its performance on predicting depth from even sparser depth samples on the \emph{NYU-Depth-v2} dataset.
\cite{ma2017sparsedense} has a similar architecture as in previous work \cite{laina2016deeper} which in turn is based on the ResNet-50 architecture introduced in \cite{he2016deep}.

Training such a network requires a vast amount of training data and obtaining suitable data with RGB and depth from real sensor data as well as ground truth depth (that is not simply inpainted) proved to be difficult.
Hence, the initial training is done on a synthetic dataset with readily available ground truth depth and publicly available RGB-D datasets are only used as validation.

\begin{figure}[t]
    \vspace{2mm}
    \centering
    \textsc{Simulation (\emph{SceneNet})}
    \includegraphics[width=0.75\columnwidth]{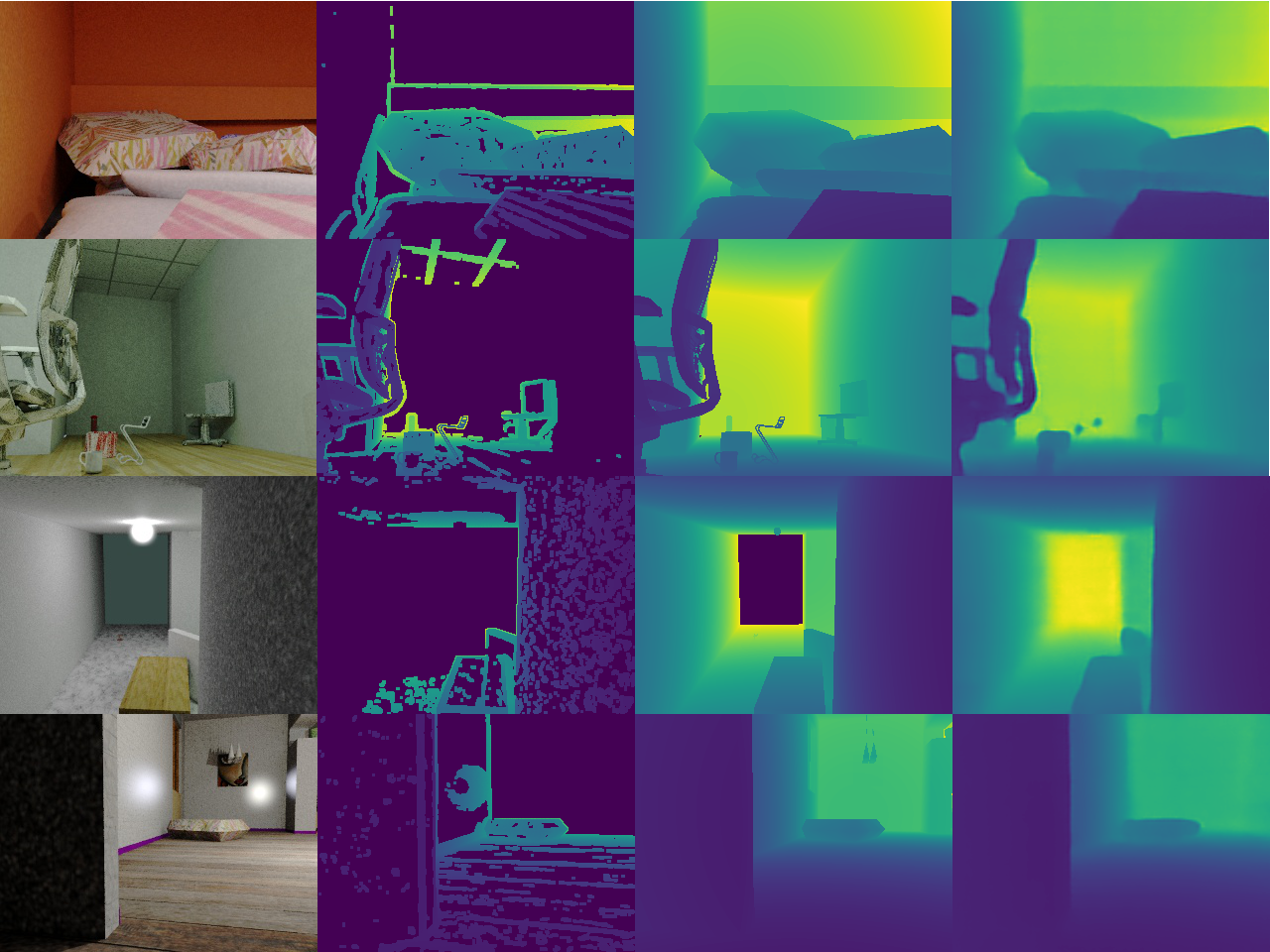}
    
    \textsc{Real Data (\emph{NYU-Depth-v2})}
    \includegraphics[width=0.75\columnwidth]{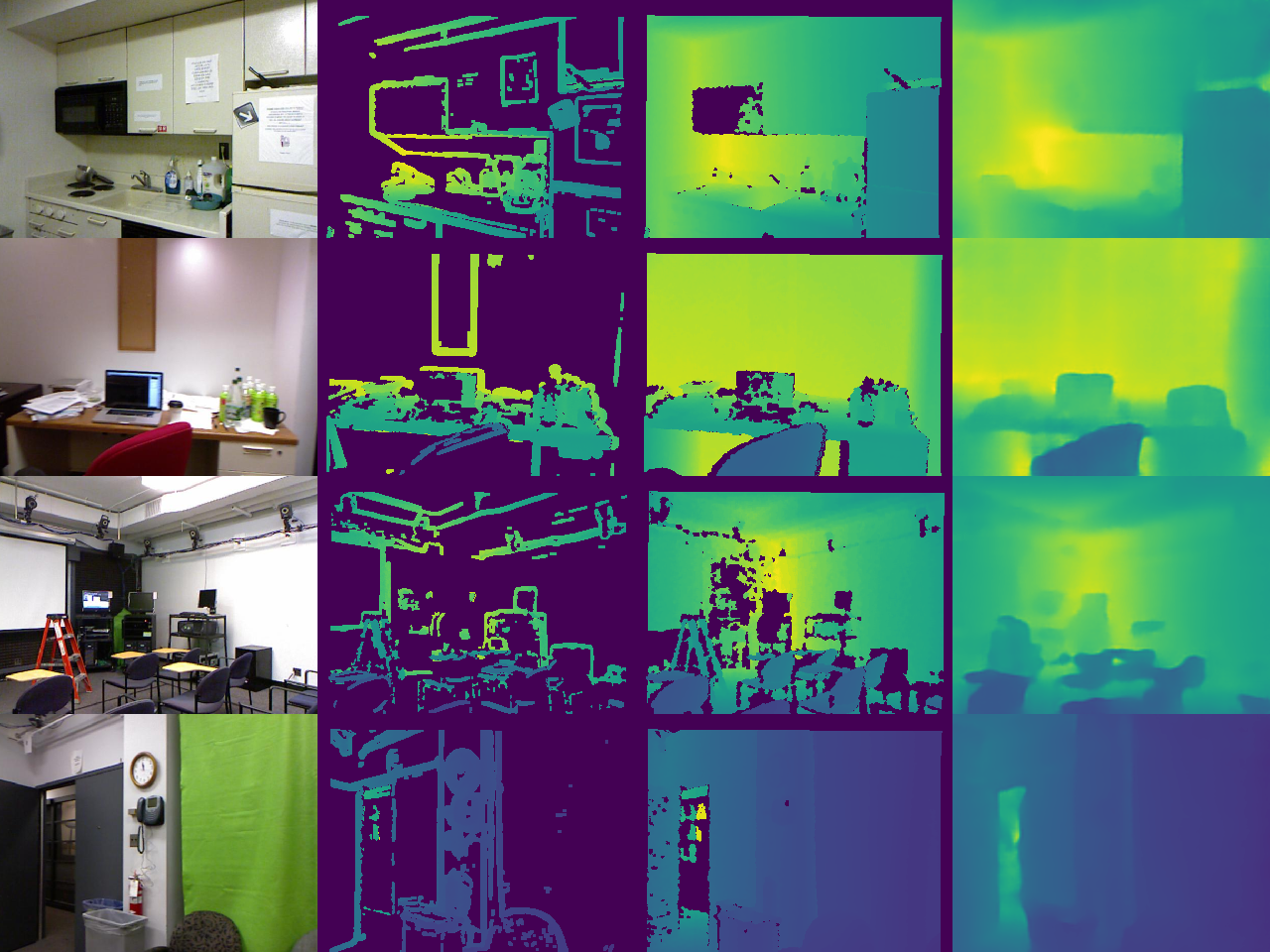}
    \caption{
    Qualitative validation on two datasets.
    Left to right: Input image, sparse depth map, ground-truth depth map and augmented depth map.
    Darkest purple indicates invalid depth.
    \emph{(top)} Four hand-picked scenes from the photorealistic \emph{SceneNet} dataset.
    Note that the window in row 3 has invalid depth in ground truth and is filled in by the depth map augmentation.
    \emph{(bottom)} Four hand-picked scenes from \emph{NYU-Depth-v2}.
    Note that the non reflective microwave in the first and the laptop in second row have partially invalid depth in the raw Kinect depth map which is completed by our model.
    }
    \label{fig::ee::dm::qual}
\end{figure}

In order to generate training data from synthetic datasets that emulates the shortcomings of a real depth sensor or semi-dense mapping techniques, we propose to apply the following sparsification to the synthetic ground truth depth maps.
The goal is to only retain parts of the depth map coinciding with areas of large image gradients as well as enforce a maximum range of the sensor.
The first part is achieved by applying a threshold to the magnitude of the edge gradients of the Gaussian-blurred image, similar to the popular canny edge detector~\cite{canny1987computational}.
The threshold is set such that a fraction $\phi$ of the depth map pixels remain.
We then apply a dilation with a $3\times3$ kernel to the mask and enforce a hard cut-off at $z_{\max} = 6m$, motivated by the report of \cite{lachat2015first} that the \emph{Kinect} sensor works well up to that range, and our own experience with similar RGB-D sensors.
Finally, for synthetic depth maps we apply noise following the noise model reported for the \emph{Kinect} sensor~\cite{nguyen2012modeling}, dropping the dependency on the observation angle $\theta$:
\begin{equation*}
    \sigma_{z}(z) = 0.0012 + 0.0019\cdot(z - 0.4)^2.
\end{equation*}
This additional noise serves to breach the reality gap and strengthen the depth augmentation performance on noisy real-world data.
Examples of this depth map sparsification can be seen in the second column of Fig.~\ref{fig::ee::dm::qual}.

It is important to note that this sparsification (except adding sensor noise) is also applied to the depth maps at inference time.
Despite throwing away potentially valuable depth measurements, it improves generalization by abstracting sensor-specific artifacts, such that depth maps from different sensors will look more alike.


\subsection{Incremental Map-Building}
\label{sec:pp:imb}
Using the camera model of the calibrated depth sensor, the augmented depth map is projected into a point cloud.
We then pass the point cloud and the corresponding pose estimate of the robot to the dense mapping framework~\cite{oleynikova2017voxblox}.
The subsequent depth fusion with the \acronym{tsdf} map is employing the widely used weighted average update function and weights the points based on their depth using a truncated quadratic weight function.
The \acronym{esdf} map used for planning is then incrementally generated and maintained based on the underlying \acronym{tsdf} map as described in~\cite{oleynikova2017voxblox}.
Please note that as proposed in~\cite{oleynikova2017voxblox} we automatically turn unknown voxels in a 3m sphere around the robot into obstacles to generate gradients in the \acronym{esdf} map that are required for the planner~\cite{oleynikovasafe2018}.
However in contrast to~\cite{oleynikova2017voxblox} when generating the global planning maps for our experiments we do not clear unknown voxels within a smaller sphere around the robot, as this would distort the results by introducing a free-space tunnel in every map completely independent of the sensor measurements.


\subsection{Planning}
\label{sec:pp:pl}
For the online planning we use the conservative planner proposed by~\cite{oleynikovasafe2018}.
At the core of the planner is a local trajectory optimizer that optimizes a high-degree polynomial spline trajectory with soft constraints on the maximum velocity and acceleration.
The intermediate goal selection used for online (re-)planning is using a similar methodology as the exploration planner next-best-view-planner proposed by~\cite{bircherreceding2016} but adds the goal to the reward function and includes several approximations to allow for ~4Hz re-planning.
For the global path planning we first obtain a initial visibility graph from start to goal position by running RRT* with an upper time limit of 5s.
We then use either~\cite{oleynikovasafe2018} or~\cite{richter2016polynomial} to obtain a dynamically feasible and collision-free trajectory.
There are some major differences between the these two methods.
\cite{oleynikovasafe2018} exploits the readily available gradients of the \acronym{esdf} map to compute an optimized trajectory that is only loosely coupled to the initial graph nodes and tries to keep a larger distance to any obstacle.
\cite{richter2016polynomial} on the other hand fits a polynomial trajectory through the waypoints of the RRT* solution and adds additional nodes only in case the fitted solution is in collision.


\section{EXPERIMENTAL SETUP}
\label{sec:es}
In this section, we describe the datasets used to train the proposed neural network as well as the evaluation metrics used in the reported results.
We then present the setup of both the simulated and real-world planning experiments.


\subsection{Datasets and Training}
\label{sec:es:dt}
For predicting complete depth maps we train a network based on the residual neural network proposed by~\cite{ma2017sparsedense}.
As input layer we use a four-channel image of size 320x240 instead of the 308x224 used in~\cite{ma2017sparsedense}.
As suggested in \cite{ma2017sparsedense} we use the mean absolute error ($\lossone$) as loss function.
To improve robustness, we augment the input data by flipping the image horizontally with $0.5$ chance and adjust brightness, contrast and saturation by factors randomly chosen between 0.6 and 1.4.
Training is performed for 10 epochs in batch sizes of 16 on a cluster equipped with GTX 1080 (TI) GPUs.
A small weight decay of $10^{-4}$ is used for regularization.
Initially the learning rate is set to 0.01 and then reduced by 20\% after 5 epochs.

To reduce the training needed, encoding layers of the network are initialized with weights pre-trained on the \emph{ImageNet}~\cite{russakovsky2015imagenet} dataset.
We then train on the synthetic \emph{SceneNet} dataset~\cite{mccormac2017scenenet} selecting around 30 images of each of the 15k indoor scenes.
%
%
The resulting dataset consists of 405k training and 24k validation pairs.

Subsequently, in order to demonstrate that the proposed training method successfully bridged the reality gap and to provide a reference to previous, comparable work, we evaluate on the \emph{NYU-Depth-v2} dataset~\cite{Silberman:ECCV12}.
This dataset, however, does not provide real ground truth, but only inpainted versions of depth maps obtained from a Kinect camera.
Therefore, we do not train on \emph{NYU-Depth-v2} but use it only for evaluation.
The raw Kinect depth maps are sparsified with our proposed method (without adding noise) and fed into the CNN.
The output of our model is then compared to the valid pixels of the raw Kinect depth map.
We evaluate the \acronym{rmse}, the \acronym{rel} and the fraction $\delta$ of pixels that lie within $\pm25\%$ of the ground truth value.

\subsection{Planning Experiments Setup}
\label{sec:es:mqm}

\begin{figure}[]
        \vspace{2mm}
	    \centering
        \includegraphics[width=.55\columnwidth]{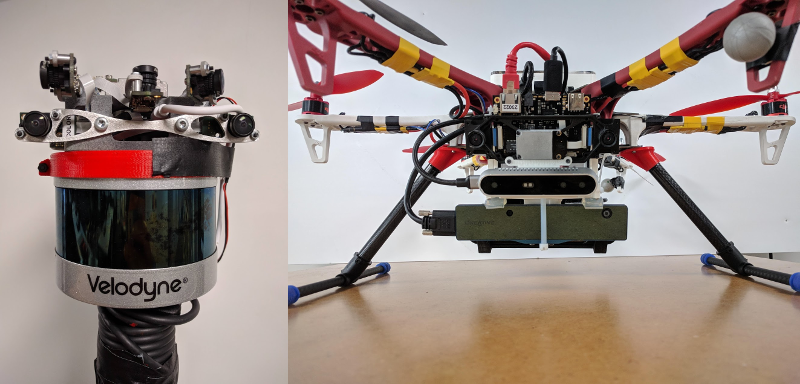}
		\caption{Visual-Inertial-RGB-D and Visual-Inertial-LiDAR sensor setup for the real world experiment.}
		\label{fig:sensors}
\end{figure}

\paragraph{Simulation} To evaluate the implications of depth augmentation on online path planning for \acronympl{mav}, we set up a virtual indoor environment in the \emph{RotorS Simulator}~\cite{Furrer2016} based on \emph{Gazebo}, consisting of four connected rooms imported from \emph{SceneNet}~\cite{mccormac2017scenenet}.
%
%
For the simulated experiments, we use the ground truth \acronym{mav} poses and color and depth images rendered at 4\,Hz.
All the processing, including depth augmentation, map building and planning, is done online and in realtime on a workstation fitted with an octacore CPU and a Nvidia GTX 1080 TI GPU.
We evaluate the planning performance (\emph{success rate}, \emph{path length}, \emph{time spent}) between $7$ hand-picked waypoints resulting in $7\cdot6 = 42$ disjoint trajectories.
A trajectory is considered \emph{successful} if a point within a maximum distance of 1\,m to the target is reached.
The path length is evaluated both in absolute, metric scale and in relation to the Euclidean distance $d = \lVert \vs - \vt \rVert_2$ between start point $\vs$ and target point $\vt$.
Path length and time spent are only evaluated on runs that are commonly successful both in \emph{ground truth} and \emph{augmented} depth.

In addition to resulting trajectories, we also evaluate the resulting \acronym{tsdf} maps from \emph{sparse} and \emph{augmented} compared to a map obtained from \emph{ground truth}.
Each voxel is labeled as either \emph{free}, \emph{occupied}, or \emph{unobserved}.
We then evaluate the number of \emph{false positive} (free voxel incorrectly classified as occupied or unobserved), \emph{false negative} (occupied voxel incorrectly classified as free).
Furthermore we evaluate the \emph{coverage}, i.e. how many voxels are observed and the \acronym{rmse} of all observed voxels.

\begin{figure}[]
        \vspace{2mm}
	    \centering
        \includegraphics[width=0.8\columnwidth]{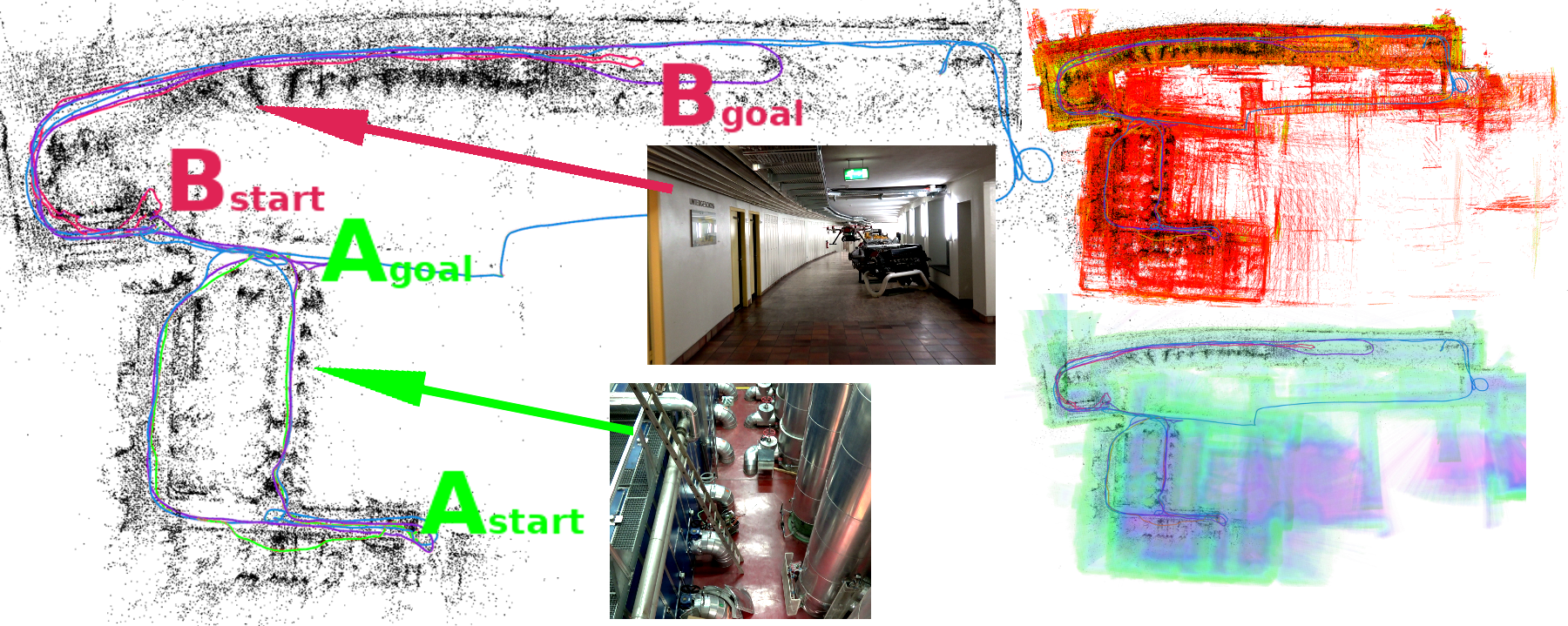}
		\caption{Maplab, a visual-inertial mapping framework~\cite{schneider2018maplab} is used to obtain consistently aligned poses of two handheld visual-inertial-LiDAR recordings (ground truth depth) and two \acronym{mav} flights. \textit{(left)} sparse pose graph, \textit{(top)} LiDAR reprojected based on optimized pose graph, \textit{(bottom)} ground truth esdf map from LiDAR.}
		\label{fig:map}
\end{figure}

\paragraph{Real world} 
In a real world \acronym{mav} experiment we manually piloted the \acronym{mav} along two trajectories in an industrial indoor environment.
The \acronym{mav} is equipped with a VI-sensor~\cite{nikolic2014synchronized} for pose estimation as well as an Intel RealSense D415 for RGB and depth measurements.
We chose the Intel Realsense D415 as a reference because based on our experience it provides one of the most dense depth maps among consumer grade sensors.
The post-processing steps applied in this sensor already represent a considerable effort from the manufacturer to obtain denser depth maps.
In order to obtain ground truth depth measurements we manually scanned the building with a Velodyne PUCK VLP-16 LiDAR attached and calibrated to another VI-sensor~\cite{honegger2014real}, see Fig.~\ref{fig:sensors}.
Using the visual-inertial mapping framework maplab~\cite{schneider2018maplab}, we computed a consistently aligned and optimized sparse feature maps based on the two \acronym{mav} flights and two handheld trajectories with the visual-inertial-LiDAR setup (see Fig.~\ref{fig:map}).
The poses of the resulting, consistent maps are used to fuse the depth measurements of both LiDAR and \acronym{mav} RGB-D sensor into different \acronym{tsdf} and subsequently \acronym{esdf} maps for planning.
In the experiment, we then evaluate the planning performance for global trajectories between start and goal of each \acronym{mav} trajectory based on the \acronym{esdf} maps obtained from the different depth sensing modalities (\emph{sparse}, \emph{raw} and \emph{augmented}).
The map obtained from LiDAR is used to obtain an upper bound (\emph{ground truth}) on the planning performance and to check if the computed paths are in fact collision free with respect to the ground truth geometry.

For all experiments we use the sparsification method described in~\autoref{sec:pp:dm} with $\phi=0.25$ and $z_{\max}=6m$.
We evaluate the path length, success rate and collisions against the current map as well as the ground truth laser map.
The planner can fail due to three reasons: if no connection to the goal is found by RRT* or the path is in collision with the map or the ground truth.
The two start and goal positions are chosen to coincide with the start and landing position of the \acronym{mav} flights.


\section{RESULTS}
\label{sec:res}

In this section, we present the results of the depth augmentation evaluation as well as the simulated and real world \acronym{mav} planning experiments.


\subsection{Depth Augmentation}
\label{sec:res:da}
\autoref{table:ee:dm:ours} compares the performance of our network to the state-of-the-art.
Please note that the degree and nature of sparsity of the input depth maps varies significantly between the different approaches.
While \cite{laina2016deeper} use color images only, \cite{liao2017parse} use a line of 225 projected LiDAR measurements and \cite{ma2017sparsedense} randomly samples 200 depth measurements.
We employ the sparsification method described in~\autoref{sec:pp:dm} and therefore the network  benefits from a larger number of depth measurements, however, their distribution highly depends on the appearance of the environment and our approach is the only one enforcing a maximum depth range.
The networks that were provided with additional depth information perform significantly better, with the proposed approach taking the lead by about $13\%$.
However, due to the significant differences of prior depth information, the result of this evaluation should not be interpreted as an advance in depth prediction performance over previous work, but merely seen as validation of our proposed training approach.

\begin{table}[b]
    \centering
    \scriptsize
    \caption{Performance of the network trained on \emph{SceneNet} compared to related work. The network has not seen any images of the \emph{NYU-Depth-v2} dataset.
    Results for \cite{laina2016deeper} are taken from  \cite{ma2017sparsedense}.}%
    \setlength\tabcolsep{3pt}
    \begin{tabular}{lllcc}
        \toprule
        Method & Dataset & Sparse Depth Input & RMSE [m] - REL - $\delta$ [\%] \\
        \midrule
        \cite{laina2016deeper} & \emph{NYUD-v2} & none (color only) & 0.573 - 0.127 - 81.1 \\
        \cite{ma2017sparsedense} & \emph{NYUD-v2} & none (color only) & 0.514 - 0.143 - 81.0 \\
        \cite{liao2017parse} & \emph{NYUD-v2} & 225 samples (line) & 0.442 - 0.104 - 87.8 \\
        \cite{ma2017sparsedense} & \emph{NYUD-v2} & 200 samples (random) & 0.230 - 0.044 - 97.1 \\
        Ours & \emph{NYUD-v2} & $\phi = 25\%, z \leq 6m$  & 0.200 - 0.043 - 98.2 \\
        Ours & \emph{SceneNet} & $\phi = 25\%, z \leq 6m$ & 0.161 - 0.053 - 98.6 \\
        \bottomrule\\
    \end{tabular}
    \label{table:ee:dm:ours}
\end{table}

As the proposed network is trained on \emph{SceneNet} it is able to perform well on the simulated scenes also based on \emph{SceneNet} used for the online planning experiments.
Considering that the network has never seen real data, the network performs well on the \emph{NYU-Depth-v2} dataset and suggests that this \acronym{cnn} is suitable for use in real world planning experiments.

%
%
%
\begin{table}[ht]
    \centering
    \caption{Planning performance of simulated and real world experiment using different depth input.}
    \scriptsize
    \centering
    \begin{center}
	\textbf{Simulation}	    
	\end{center} 
	\begin{tabular}{l r r r}
		\toprule
		Sensor & Success (\# - rate) & Path (abs - rel) & Time \\
		\midrule
		\emph{ground truth} & 17/42 - 40\% & 12.46m - $1.25\cdot d$ & 22.55s \\
 		\emph{augmented} & 16/42 - 38\% & 12.51m - $1.26\cdot d$ & 23.29s \\
		\emph{sparse} & 7/42 - 17\% & 12.91m - $1.29\cdot d$ & 22.94s \\
		\bottomrule\\
	\end{tabular}
	\begin{center}
	\textbf{Real World Experiment}
	\end{center}
	\setlength\tabcolsep{2.5pt}
    \resizebox{\columnwidth}{!}{%
    \begin{tabular}{lc|c|c|c|}
     \multicolumn{5}{@{}p{\columnwidth}@{}}{%
   	Global planning between the two waypoints pairs A and B, corresponding to start and goal of the two \acronym{mav} flights (See Fig.~\ref{fig:map}). The results are reported for 300 trials for each set of waypoints.
   	R: Radius of collision check [m], S: Success in [\%], C: Failure due to collision with ground truth [\%], L: Average path length [m]}\\
   	\multicolumn{5}{c}{} \\
    \multicolumn{5}{c}{Waypoints A} \\
    \toprule
     &  & R = 0.25 & R = 0.4 & R = 0.5 \\
    Sensor & Planner & S / C / L & S / C / L & S / C / L \\
    \midrule
    \multirow{2}{*}{LiDAR} & \cite{richter2016polynomial}                                                         & 91.7          / ---           / 18.1 &   100.0          / ---           / 19.0  &   100.0          / ---           / 19.2 \\
     & \cite{oleynikovasafe2018}                                                                               & 61.7          / ---           / 18.2 &    97.0          / ---           / 19.3  &    96.0          / ---           / 19.4 \vspace{1mm} \\
    \multirow{2}{*}{Ours} & \cite{richter2016polynomial}                                                          &  \textbf{8.7} / 86.7          / 18.4 &    \textbf{23.0} / \textbf{72.0} / 18.9  &    \textbf{40.3} / \textbf{56.3} / 20.8 \\
     & \cite{oleynikovasafe2018}                                                                               & \textbf{45.3} / \textbf{49.0} / 19.2 &    \textbf{98.3} /  \textbf{0.7} / 19.4  &    83.7          /  \textbf{1.7} / 20.4 \vspace{1mm} \\
    \multirow{2}{*}{\begin{tabular}[c]{@{}l@{}}Raw\\ (\textless 10m)\end{tabular}} & \cite{richter2016polynomial} &  0.0          / \textbf{79.3} / ---  &     0.0          / 92.0          / ---   &    12.3          / 83.0          / 19.0 \\
     & \cite{oleynikovasafe2018}                                                                               &  0.3          / 86.3          / 19.5 &    29.0          / 44.3          / 19.1  &    \textbf{88.3} / 10.0          / 19.6 \vspace{1mm} \\
    \multirow{2}{*}{Raw} & \cite{richter2016polynomial}                                                           &  0.3          / 90.7          / 19.9 &     6.7          / 80.3          / 19.2  &     9.3          / 80.3          / 20.8 \\
     & \cite{oleynikovasafe2018}                                                                               & 11.3          / 83.3          / 18.5 &    65.0          / 26.0          / 19.1  &    55.3          / 35.3          / 20.3 \vspace{1mm} \\
    Sparse & \cite{richter2016polynomial} / \cite{oleynikovasafe2018}                                          & ---                                  & ---                                      & --- \\
     \bottomrule
     \multicolumn{5}{c}{} \\
     \multicolumn{5}{c}{Waypoints B} \\
     \toprule
      &  & R = 0.25 & R = 0.4 & R = 0.5 \\
     Sensor & Planner & S / C / L & S / C / L & S / C / L \\
    \midrule
    \multirow{2}{*}{LiDAR} & \cite{richter2016polynomial}                                                         & 94.7          / ---           / 29.1 & 97.0          / ---           / 29.6 & 94.7          / ---           / 29.9 \\
     & \cite{oleynikovasafe2018}                                                                               & 99.0          / ---           / 29.9 & 97.7          / ---           / 30.3 & 98.3          / ---           / 30.7  \vspace{1mm}\\
    \multirow{2}{*}{Ours} & \cite{richter2016polynomial}                                                          & 78.7          / 9.3           / 38.7 & \textbf{73.3} / \textbf{6.3}  / 39.4 & \textbf{66.3} / \textbf{15.7} / 40.3 \\
     & \cite{oleynikovasafe2018}                                                                               & \textbf{66.0} / 10.0          / 41.4 & \textbf{95.7} / \textbf{0.3}  / 42.4 & 56.7          / \textbf{0.3}  / 43.7  \vspace{1mm}\\
    \multirow{2}{*}{\begin{tabular}[c]{@{}l@{}}Raw\\ (\textless 10m)\end{tabular}} & \cite{richter2016polynomial} & \textbf{94.3} /  \textbf{2.0} / 38.1 & 21.7          / 10.0          / 39.4 & 50.0          / 43.0          / 42.4 \\
     & \cite{oleynikovasafe2018}                                                                               & 66.7          /  7.3          / 40.3 & 50.7          / 0.7           / 42.1 & \textbf{61.7} /  9.7          / 45.6  \vspace{1mm}\\
    \multirow{2}{*}{Raw} & \cite{richter2016polynomial}                                                           &  0.0          / 95.3          / ---  &  0.0          / 86.0          / ---  &  6.7          / 54.0          / 42.0 \\
     & \cite{oleynikovasafe2018}                                                                               &  0.0          / 69.7          / ---  &  0.0          / 75.3          / ---  &  5.8          / 21.3          / 44.4  \vspace{1mm}\\
    \multirow{2}{*}{Sparse} & \cite{richter2016polynomial}                                                        & 71.0          / 22.3          / 41.5 & ---                                  & --- \\
      & \cite{oleynikovasafe2018}                                                                              & 31.7          /  \textbf{0.6} / 44.1 & ---                                  & --- \\
     \bottomrule
    \end{tabular}
    }
\label{table:plan:results}
\end{table}

\begin{table}[ht]
    \scriptsize
    \centering
    \caption{
    Errors of the TSDF from sparse respectively augmented depths compared to TSDF obtained from simulation ground truth.}
    \vspace{-0.3cm}
    \begin{center}
	\textbf{Simulation}	    
	\end{center} 
    \begin{tabular}{lccrr}
        \toprule
        Method & False Pos.\ & False Neg.\ & Coverage & RMSE  \\
        \midrule
        \emph{Sparse} & 35\% & 21.7\% & 89\% & 35mm \\
        \emph{Augmented} & 0.73\% & 28.8\% & 99\%  & 39mm \\
        \bottomrule\\
    \end{tabular}
	\vspace{-0.3cm}
	%
    \label{table:ee:blox}
\end{table}
\begin{figure}[ht]
    \vspace{2mm}
    \centering
    \includegraphics[width=\columnwidth]{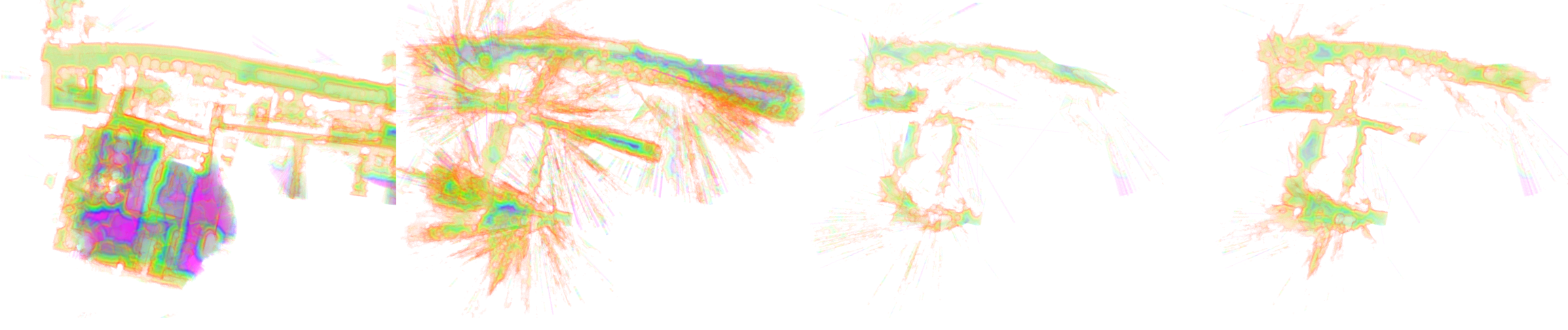}
    \caption{Visualization of the traversability of the \acronym{esdf} map, computed by~\cite{oleynikova2017voxblox}. From left to right: \emph{lidar}, \emph{raw}, \emph{sparse} and \emph{augmented depth}.} 
    \label{fig:re:trav}
\end{figure}
\subsection{Online Planning in Simulation}
\label{sec:res:pe}
Results from simulated planning experiments are summarized in \autoref{table:plan:results} and five exemplary trajectories of successful runs using \emph{augmented} depths are plotted in Fig.~\ref{fig:traj}.
Using \emph{sparse} depth only 7 runs were successful whereas using \emph{augmented} depth this number is doubled and approaches the success rate of \emph{ground truth}.
The planning performance is almost on par to \emph{ground truth} in both \emph{path length} and \emph{time spent}.
It should be noted that the absolute metrics in \emph{sparse} are averaged over only seven paths.
These paths tend to be between nearby waypoints and hence the relative \emph{path length} should be compared.
The longer path length is due to uncleared space blocking more direct ways resulting in detours (e.g. in Fig.~\ref{fig:traj}).

%
%
%
%
%
\subsection{Global Planning on Real World Data} 
\autoref{table:plan:results} summarizes the results of the real world planning experiment.
We observed that the two sets of waypoints (A and B) represented two very different planning scenarios, see Fig.~\ref{fig:map}.
The lower, industrial floor (A) has both narrow and very open sections, challenging materials and shapes, leaving the planners with a significant amount of unobserved space and thin structures (e.g. a ladder) to avoid.
We observe that especially for the smaller collision radius,~\cite{richter2016polynomial} struggles to find a path for all maps, as it closely follows the more risky initial path provided by RRT*, which leads to collisions with the ground truth map due to map inaccuracies (most likely missing thin structures).
In this challenging environment, the augmented sensor performs significantly better than all other sensors with higher success rates and lower collision rates.

The upper, wide, well-textured corridor environment (B) allows for successful planning even based on the \emph{sparse} depth sensor, which surprisingly even surpasses the \emph{raw} depth sensor.
Upon closer inspection of this we noticed that due to erroneous depth measurements at large distance the planner tried plan through the outer wall of the upper floor instead of following the stair case.
Hence, we also evaluated based on the depth with large depth values removed (\emph{raw} ($< 10m$)).
For B the \emph{augmented} sensor still performs slightly better than the filtered \emph{raw} sensor however the difference is not as pronounced as for~A.

The results of the \emph{augmented} planning system on the real data exceeded our expectations, especially considering it has never seen real data before.
Overall \emph{augmented} planning system reaches a better success rate than the planner using the filtered \emph{raw} depth, and clearly outperforms the planners based on \emph{sparse} and unfiltered \emph{raw} depth.
Another significant and somewhat surprising results is that overall the \emph{augmented} sensor resulted in less collisions with the ground truth map which would suggest that at least in this particular scenario it is safer.
The evaluation of the path length did not reveal any significant difference between the sensor modalities aside from the fact that for B, the LiDAR sensor enabled the planner to take a shortcut due to its superior range and field of view, see Fig.~\ref{fig:re:trav}.

\section{CONCLUSION}
\label{sec:c}
This work proposes an \emph{augmented} planning system, evaluating the impact of employing a residual neural network to improve depth perception for planning based on sparse/semi-dense depth sensor data.
The depth augmentation network was trained solely on simulated noisy and sparse depth data and achieved state-of-the-art performance in depth augmentation when evaluated on real data.
The complete \emph{augmented} system, from perception to path planning was then evaluated in simulated and real \acronym{mav} planning experiments.
The \emph{augmented} system not only demonstrated near ground truth performance in a simulated online planning experiment, but also showed promising results in the real world \acronym{mav} experiment, outperforming a very dense, consumer grade RGB-D sensor, despite having access to only sparse/semi-dense depth sensing.
Unsurprisingly the best planning performance is achieved in combination with a conservative planner that favors trajectories that are further away from the obstacles and therefore more robust against inaccuracies of the depth prediction.
To the best of our knowledge, this is the first system employing depth augmentation in a practical \acronym{mav} path planning pipeline, tackling real problems with depth perception.
One major limitation of the system is the lack of a confidence measure for the predicted depth, which would allow for a more advanced fusion of predicted and measured depth and ultimately a safer planning pipeline.
Another limitation we observed in our evaluation is that using the whole raw depth information for prediction does not improve but significantly degrade the depth augmentation performance (\acronym{rmse}: 0.34m vs.\ 0.20m for sparse input).
We believe that in order to overcome this over-fitting to specific type of sparsity, a large quantity and variety of real sensor data with accurate ground truth is required.


\section*{ACKNOWLEDGMENT}
We would like to thank Helen Oleynikova for her help with the planner, Zachary Taylor for enabling the real world experiments and providing the VI-LiDAR setup and Fangchang Ma for his help adapting his network.

\addtolength{\textheight}{-2cm}

\bibliographystyle{IEEEtran} 
\small
\bibliography{main}  

\begin{thebibliography}{10}
\providecommand{\url}[1]{#1}
\csname url@rmstyle\endcsname
\providecommand{\newblock}{\relax}
\providecommand{\bibinfo}[2]{#2}
\providecommand\BIBentrySTDinterwordspacing{\spaceskip=0pt\relax}
\providecommand\BIBentryALTinterwordstretchfactor{4}
\providecommand\BIBentryALTinterwordspacing{\spaceskip=\fontdimen2\font plus
\BIBentryALTinterwordstretchfactor\fontdimen3\font minus
  \fontdimen4\font\relax}
\providecommand\BIBforeignlanguage[2]{{%
\expandafter\ifx\csname l@#1\endcsname\relax
\typeout{** WARNING: IEEEtran.bst: No hyphenation pattern has been}%
\typeout{** loaded for the language `#1'. Using the pattern for}%
\typeout{** the default language instead.}%
\else
\language=\csname l@#1\endcsname
\fi
#2}}

\bibitem{pivtoraiko2013incremental}
M.~Pivtoraiko, D.~Mellinger, and V.~Kumar, ``Incremental micro-uav motion
  replanning for exploring unknown environments,'' in \emph{ICRA}.\hskip 1em
  plus 0.5em minus 0.4em\relax IEEE, 2013, pp. 2452--2458.

\bibitem{chen2016online}
J.~Chen, T.~Liu, and S.~Shen, ``Online generation of collision-free
  trajectories for quadrotor flight in unknown cluttered environments,'' in
  \emph{ICRA}.\hskip 1em plus 0.5em minus 0.4em\relax IEEE, 2016, pp.
  1476--1483.

\bibitem{oleynikovasafe2018}
H.~Oleynikova, Z.~Taylor, R.~Siegwart, and J.~Nieto, ``Safe local exploration
  for replanning in cluttered unknown environments for microaerial vehicles,''
  \emph{RA-L}, vol.~3, no.~3, pp. 1474--1481, 2018.

\bibitem{ma2017sparsedense}
F.~Ma and S.~Karaman, ``Sparse-to-dense: Depth prediction from sparse depth
  samples and a single image,'' in \emph{ICRA}.\hskip 1em plus 0.5em minus
  0.4em\relax IEEE, 2018, pp. 1--8.

\bibitem{oleynikova2017voxblox}
H.~Oleynikova, Z.~Taylor, M.~Fehr, R.~Siegwart, and J.~Nieto, ``Voxblox:
  Incremental 3d euclidean signed distance fields for on-board mav planning,''
  in \emph{IROS}.\hskip 1em plus 0.5em minus 0.4em\relax IEEE, 2017, pp.
  1366--1373.

\bibitem{richter2016polynomial}
C.~Richter, A.~Bry, and N.~Roy, ``Polynomial trajectory planning for aggressive
  quadrotor flight in dense indoor environments,'' in \emph{Robotics
  Research}.\hskip 1em plus 0.5em minus 0.4em\relax Springer, 2016, pp.
  649--666.

\bibitem{liu2015deep}
F.~Liu, C.~Shen, and G.~Lin, ``Deep convolutional neural fields for depth
  estimation from a single image,'' in \emph{CVPR}, 2015, pp. 5162--5170.

\bibitem{eigen2015predicting}
D.~Eigen and R.~Fergus, ``Predicting depth, surface normals and semantic labels
  with a common multi-scale convolutional architecture,'' in \emph{ICCV}, 2015,
  pp. 2650--2658.

\bibitem{laina2016deeper}
I.~Laina, C.~Rupprecht, V.~Belagiannis, F.~Tombari, and N.~Navab, ``Deeper
  depth prediction with fully convolutional residual networks,'' in
  \emph{3DV}.\hskip 1em plus 0.5em minus 0.4em\relax IEEE, 2016, pp. 239--248.

\bibitem{roy2016monocular}
A.~Roy and S.~Todorovic, ``Monocular depth estimation using neural regression
  forest,'' in \emph{CVPR}, 2016, pp. 5506--5514.

\bibitem{pan2018depth}
L.~Pan, Y.~Dai, M.~Liu, and F.~Porikli, ``Depth map completion by jointly
  exploiting blurry color images and sparse depth maps,'' in \emph{WACV}.\hskip
  1em plus 0.5em minus 0.4em\relax IEEE, 2018, pp. 1377--1386.

\bibitem{liao2017parse}
Y.~Liao, L.~Huang, Y.~Wang, S.~Kodagoda, Y.~Yu, and Y.~Liu, ``Parse geometry
  from a line: Monocular depth estimation with partial laser observation,'' in
  \emph{ICRA}.\hskip 1em plus 0.5em minus 0.4em\relax IEEE, 2017, pp.
  5059--5066.

\bibitem{martins2018fusion}
D.~Martins, K.~Van~Hecke, and G.~De~Croon, ``Fusion of stereo and still
  monocular depth estimates in a self-supervised learning context,'' in
  \emph{ICRA}.\hskip 1em plus 0.5em minus 0.4em\relax IEEE, 2018, pp. 849--856.

\bibitem{jaritz2018sparse}
M.~Jaritz, R.~De~Charette, E.~Wirbel, X.~Perrotton, and F.~Nashashibi, ``Sparse
  and dense data with cnns: Depth completion and semantic segmentation,'' in
  \emph{3DV}.\hskip 1em plus 0.5em minus 0.4em\relax IEEE, 2018, pp. 52--60.

\bibitem{pilzer2018unsupervised}
A.~Pilzer, D.~Xu, M.~Puscas, E.~Ricci, and N.~Sebe, ``Unsupervised adversarial
  depth estimation using cycled generative networks,'' in \emph{3DV}.\hskip 1em
  plus 0.5em minus 0.4em\relax IEEE, 2018, pp. 587--595.

\bibitem{zhang2018deep}
Y.~Zhang and T.~Funkhouser, ``Deep depth completion of a single rgb-d image,''
  in \emph{CVPR}, 2018, pp. 175--185.

\bibitem{cheng2018depth}
X.~Cheng, P.~Wang, and R.~Yang, ``Depth estimation via affinity learned with
  convolutional spatial propagation network,'' in \emph{ECCV}, 2018, pp.
  103--119.

\bibitem{ma2018self}
F.~Ma, G.~V. Cavalheiro, and S.~Karaman, ``Self-supervised sparse-to-dense:
  self-supervised depth completion from lidar and monocular camera,'' in
  \emph{ICRA}.\hskip 1em plus 0.5em minus 0.4em\relax IEEE, 2019, pp.
  3288--3295.

\bibitem{imran2019depth}
S.~Imran, Y.~Long, X.~Liu, and D.~Morris, ``Depth coefficients for depth
  completion,'' \emph{arXiv preprint}, 2019.

\bibitem{yang20173d}
B.~Yang, H.~Wen, S.~Wang, R.~Clark, A.~Markham, and N.~Trigoni, ``3d object
  reconstruction from a single depth view with adversarial learning,'' in
  \emph{ICCV}, 2017, pp. 679--688.

\bibitem{dai2017scancomplete}
A.~Dai, D.~Ritchie, M.~Bokeloh, S.~Reed, J.~Sturm, and M.~Nie{\ss}ner,
  ``Scancomplete: Large-scale scene completion and semantic segmentation for 3d
  scans,'' in \emph{CVPR}, 2017.

\bibitem{kim2015deep}
D.~K. Kim and T.~Chen, ``Deep neural network for real-time autonomous indoor
  navigation,'' \emph{arXiv preprint}, 2015.

\bibitem{ross2013learning}
S.~Ross, N.~Melik-Barkhudarov, K.~S. Shankar, A.~Wendel, D.~Dey, J.~A. Bagnell,
  and M.~Hebert, ``Learning monocular reactive uav control in cluttered natural
  environments,'' in \emph{ICRA}.\hskip 1em plus 0.5em minus 0.4em\relax IEEE,
  2013, pp. 1765--1772.

\bibitem{loquercio2018dronet}
A.~Loquercio, A.~I. Maqueda, C.~R. del Blanco, and D.~Scaramuzza, ``Dronet:
  Learning to fly by driving,'' \emph{RA-L}, vol.~3, no.~2, pp. 1088--1095,
  2018.

\bibitem{richtersafe2017}
C.~Richter and N.~Roy, ``Safe visual navigation via deep learning and novelty
  detection,'' in \emph{RSS}, 2017.

\bibitem{he2016deep}
K.~He, X.~Zhang, S.~Ren, and J.~Sun, ``Deep residual learning for image
  recognition,'' in \emph{CVPR}, 2016, pp. 770--778.

\bibitem{canny1987computational}
J.~Canny, ``A computational approach to edge detection,'' in \emph{Readings in
  Computer Vision}.\hskip 1em plus 0.5em minus 0.4em\relax Elsevier, 1987, pp.
  184--203.

\bibitem{lachat2015first}
E.~Lachat, H.~Macher, M.~Mittet, T.~Landes, and P.~Grussenmeyer, ``First
  experiences with kinect v2 sensor for close range 3d modelling,'' \emph{ISPRS
  Archives}, vol.~40, no.~5, p.~93, 2015.

\bibitem{nguyen2012modeling}
C.~V. Nguyen, S.~Izadi, and D.~Lovell, ``Modeling kinect sensor noise for
  improved 3d reconstruction and tracking,'' in \emph{3DIMPVT}.\hskip 1em plus
  0.5em minus 0.4em\relax IEEE, 2012, pp. 524--530.

\bibitem{bircherreceding2016}
A.~Bircher, M.~Kamel, K.~Alexis, H.~Oleynikova, and R.~Siegwart, ``Receding
  horizon" next-best-view" planner for 3d exploration,'' in \emph{ICRA}.\hskip
  1em plus 0.5em minus 0.4em\relax IEEE, 2016, pp. 1462--1468.

\bibitem{russakovsky2015imagenet}
O.~Russakovsky, J.~Deng, H.~Su, J.~Krause, S.~Satheesh, S.~Ma, Z.~Huang,
  A.~Karpathy, A.~Khosla, M.~Bernstein, \emph{et~al.}, ``Imagenet large scale
  visual recognition challenge,'' \emph{International journal of computer
  vision}, vol. 115, no.~3, pp. 211--252, 2015.

\bibitem{mccormac2017scenenet}
J.~McCormac, A.~Handa, S.~Leutenegger, and A.~J. Davison, ``Scenenet rgb-d: Can
  5m synthetic images beat generic imagenet pre-training on indoor
  segmentation,'' in \emph{ICCV}, vol.~1, 2017.

\bibitem{Silberman:ECCV12}
P.~K. Nathan~Silberman, Derek~Hoiem and R.~Fergus, ``Indoor segmentation and
  support inference from rgbd images,'' in \emph{ECCV}, 2012.

\bibitem{Furrer2016}
F.~Furrer, M.~Burri, M.~Achtelik, and R.~Siegwart, \emph{Robot Operating System
  (ROS): The Complete Reference (Volume 1)}.\hskip 1em plus 0.5em minus
  0.4em\relax Cham: Springer International Publishing, 2016, ch. RotorS---A
  Modular Gazebo MAV Simulator Framework, pp. 595--625.

\bibitem{nikolic2014synchronized}
J.~Nikolic, J.~Rehder, M.~Burri, P.~Gohl, S.~Leutenegger, P.~T. Furgale, and
  R.~Siegwart, ``A synchronized visual-inertial sensor system with fpga
  pre-processing for accurate real-time slam,'' in \emph{ICRA}.\hskip 1em plus
  0.5em minus 0.4em\relax IEEE, 2014, pp. 431--437.

\bibitem{honegger2014real}
D.~Honegger, H.~Oleynikova, and M.~Pollefeys, ``Real-time and low latency
  embedded computer vision hardware based on a combination of fpga and mobile
  cpu,'' in \emph{IROS}.\hskip 1em plus 0.5em minus 0.4em\relax IEEE, 2014, pp.
  4930--4935.

\bibitem{schneider2018maplab}
T.~Schneider, M.~Dymczyk, M.~Fehr, K.~Egger, S.~Lynen, I.~Gilitschenski, and
  R.~Siegwart, ``maplab: An open framework for research in visual-inertial
  mapping and localization,'' \emph{RA-L}, vol.~3, no.~3, pp. 1418--1425, 2018.

\end{thebibliography}

\end{document}